\documentclass{article}
\usepackage{amsmath}
\usepackage{caption}
\usepackage{indentfirst}
\usepackage{booktabs}
\usepackage{multirow}
\usepackage{float}
\newtheorem{rmk}{Remark}

\usepackage{PRIMEarxiv}
\usepackage[utf8]{inputenc} 
\usepackage[T1]{fontenc}    
\usepackage{hyperref}       
\usepackage{url}            
\usepackage{booktabs}       
\usepackage{amsfonts}       
\usepackage{nicefrac}       
\usepackage{microtype}      
\usepackage{lipsum}
\usepackage{fancyhdr}       
\usepackage{graphicx}       
\graphicspath{{media/}}     

\title{Exploring Different Time-series-Transformer (TST) Architectures: A Case Study in Battery Life Prediction for Electric Vehicles (EVs)
}

\author{
  Niranjan Sitapure$^{*}$ \\
  Dept. of Chemical Engineering \\
  Texas A\&M University \\
  College Station, TX 77801\\
  \texttt{niranjan\_sitapure@tamu.edu} \\
   \And
  Atharva Kulkarni\\
  Dept. of Mechanical Engineering \\
  Texas A\&M University\\
  College Station, TX 77801\\
  \texttt{atharva1609@tamu.edu} \\
}
\begin{document}

\captionsetup[figure]{font=small,skip=10pt}
\captionsetup[table]{font=small,skip=10pt}

\maketitle

\begin{abstract}
In the past few years, the development of battery technology for application in electric vehicles (EVs) has received significant attention. Although a major part of the efforts has been geared  towards the development of new battery materials and chemistries, another challenge faced by EVs is an accurate prediction of key battery parameters (e.g., state-of-charge (SOC), temperature, and others), which are essential in the construction of advanced battery management systems (BMS). Despite the presence of a plethora of battery models (e.g., equivalent circuit model, single-particle model, etc.), all the parameters that affect battery parameters cannot be covered by these models in a computationally tractable manner. Specifically, EV operation includes various non-battery-related parameters (e.g., ambient temperature, user-defined cabin temperature, elevation, regenerative braking, etc.) that affect battery performance. Since the incorporation of these auxiliary parameters in traditional battery models is difficult, a data-driven approach is suggested to capture their effect on battery performance. More precisely, given the emergence of time-series-transformers (TSTs) that harness the power of multiheaded attention and parallelization-friendly architecture, typical  state-of-the-art (SOTA) TST models (e.g., encoder-only, and vanilla-TST) are explored and compared against LSTM. Furthermore, novel and unseen TST architectures (i.e., encoder TST + decoder LSTM, and a hybrid TST-LSTM) are also developed and compared against the aforementioned models. Also, to address the challenge of battery life prediction in EVs using these novel TST architectures, a dataset comprising 72 unique driving trips performed in a BMW i3 (60 Ah) is considered. Here the objective is to develop accurate TST models that take in environmental data, battery data, vehicle driving data, and heating circuit data to predict SOC and battery temperature for future time steps (i.e., 1 min to 5 mins).

\end{abstract}

\section{Introduction}
\setlength{\parindent}{15pt} 

In recent years, the focus on advancing battery technology for electric vehicles (EVs) has intensified, reflecting the industry's commitment to enhancing EV performance and sustainability \cite{wang_review_2018, nitta2015li, nykvist2015rapidly}. Extensive research and development efforts have been dedicated to creating innovative battery materials and chemistries, aiming to optimize the overall efficiency and longevity of EV batteries \cite{parekh2022critical, manthiram2020reflection, zhao2021review}. Despite these advancements, a critical challenge remains in accurately predicting key battery parameters, such as state-of-charge (SOC) and temperature, which are pivotal in the design and implementation of sophisticated battery management systems (BMS) \cite{sitapure2020computational, lee2021multiscale, pozzi2020optimal}. While various battery models, such as the equivalent circuit model and single-particle model, have been formulated, they often fall short of providing a comprehensive representation of all factors that significantly influence battery performance. \cite{hwang2022model, torchio_real-time_2015, suthar_optimal_2013}. These models struggle to strike the right balance between computational efficiency and encompassing the multitude of variables that impact EV battery behavior. Among these factors are non-battery-related parameters, including ambient temperature, user-defined cabin temperature, elevation changes, and regenerative braking during EV operation \cite{dehghani2019study, keil2015aging, steinstraeter2020range}. These external factors can exert a noticeable influence on battery performance, but their integration into traditional battery models proves challenging due to the complexities involved. Thus, given the massive advent of fast and accurate machine learning (ML) and artificial intelligence (AI) tools, a data-driven approach toward real-time battery life prediction is a priority for various EV manufacturers and academics. \cite{tian2017understanding, lv2022machine,  severson2019data, ng2020predicting}. 

Within the field of data-driven prediction of battery performance, there is a platter of variegated approaches. For instance, Kwon and colleagues have extensively explored various subspace identification techniques like sparse identification of system dynamics (SINDy) and operable adaptive sparse identification of systems (OASIS) \cite{bhadriraju2019machine, Bhadriraju2019, bhadriraju2021oasis}. Recently, the authors showcased the ability of a two-timescale OASIS model to handle intra-cycle and inter-cycle battery degradation to accurately predict SOC in lithium batteries \cite{bhadriraju2023adaptive}.  On the other hand, Braatz and colleagues developed an ML approach that employs a feature-based approach, where linear or nonlinear transformations of raw data are generated and used within a regularized linear framework called the elastic net. The model predicts the logarithm of cycle life by using a linear combination of selected features \cite{severson2019data}. More recently, Kim and colleagues developed a long-short-term-memory (LSTM)-based prediction of remaining battery life using a multichannel array of battery parameters \cite{park2020lstm}. That being said, most of the above models, utilize various battery parameters (e.g., voltage, current, temperature, etc.) to predict SOC. However, during EV operation various environmental parameters (e.g., ambient temperature, elevation change, etc.), vehicle driving data (e.g., regenerative braking, velocity, acceleration, etc), and heating circuit conditions (e.g., cabin temperature, air vent temperature, etc.) also affect battery SOC. Thus, an adept ML model that can assign adaptive weights to these factors in addition to battery parameters will provide an accurate prediction of real-time SOC during the operation of an EV. 

To this end, it is worthwhile to explore the emergence of transformer models that are used to construct remarkable powerful tools like ChatGPT, Bard, CodeGPT, and many others \cite{GPT4_technical_report, devlin2018bert,vaswani2017attention,radford2019language,brown2020language}.  These models have showcased their capabilities in a wide range of tasks, from generating human-like text responses to understanding and generating programming code, mainly due to their utilization of several key mechanisms: multiheaded attention, positional encoding (PE), and a parallelization-friendly architecture. The attention mechanism is a fundamental component of transformer models that enables the models to effectively capture and weigh the relationships between different parts of the input data \cite{devlin2018bert}. By conducting scaled-dot product calculations between various input tensors, the attention mechanism selectively assigns higher attention scores to time-steps that exhibit significant process changes or specific patterns, allowing the models to focus on relevant information while processing data \cite{radford2019language}. Also, PE plays a crucial role as these models do not inherently encode the sequential nature of input data (unlike LSTM models), positional encoding provides a systematic way to incorporate sequence information \cite{vaswani2017attention}. This allows the models to understand the relative positions of elements in the input sequence, preserving the temporal context and aiding in the recognition of patterns across time. Finally, the parallelization-friendly architecture enables the generation of very large transformer models (i.e., model parameters ($N_p \in [1, 1000]$ M) that process a humongous dataset that has been compiled from a plethora of different sources, thereby enabling a high degree of multidisciplinary transfer learning \cite{shoeybi2019megatron}. Although the initial development of transformer models was restricted to large-language-models (LLMs), there have been several recent advances in the development of time-series-transformers (TSTs) that are applicable to a wide variety of chemical, mechanical, and electrical engineering applications \cite{sitapure2023exploring, wen2022transformers}. Most notably, Kwon and Sitapure showcased \textit{CrystalGPT}, which is a unified digital twin for 20+ different sugar crystal systems, and it is approximately 10 times more accurate than other state-of-the-art (SOTA) ML methods \cite{sitapure2023crystalgpt}. Similarly, Pfister and colleagues have tested multi-horizon temporal fusion transformers on various real-world datasets including power-consumption predictions by residents, volatility in the stock market, and consumer spending in retail stores \cite{lim2021temporal}. Moreover, TSTs also utilize attention mechanism and leverage process data (e.g., temperature, concentration, and crystal size in the case of \textit{CrystalGPT}) or market data (e.g., stock indices for 50+ volatile stocks and 100+ indicators for big-cap companies, etc.) for the current and preceding $W$ time-steps (with a window size of $W$) to predict the features over next $H$ time-steps. This approach facilitates a contextual understanding of both short-term and long-term changes in process states, providing valuable insights into the overall process dynamics. 

The above demonstrations indicate that TSTs can potentially be a promising avenue for developing data-driven models for battery life predictions in EVs. Specifically, typical TST architectures (e.g., encoder-only, and vanilla-TST) were developed and compared with LSTM models, which are considered to be incumbents for time-series predictions. Furthermore, novel and unseen hybrid-TST architectures were also constructed. For instance, a hybrid-TST model with an encoder part pertaining to a typical TST network is combined with a decoder comprising of simply multiple layers of an LSTM network is developed. Similarly, another hybrid-TST model that utilizes the same architecture as a vanilla-TST, albeit with replacing the feed-forward network (FFN) with an LSTM layer is developed. The rationale here is that the attention mechanism empowers the TST model to adaptively weigh different features and learn their underlying interdependencies while the LSTM is adept at time-series modeling, and thereby combining these two aspects can potentially outperform existing TST and LSTM models. To evaluate this hypothesis, we utilize a dataset comprising 72 unique driving tests performed on a BMW i3 (60 Ah) \cite{bmw_dataset}. This dataset includes environmental data, battery data, vehicle driving data, and heating circuit data. The primary objective is to create TST models that can effectively predict SOC and battery temperature for future time-steps, ranging from 1 minute to 5 minutes. Overall, the current work addresses two challenges; (a) the development of fast and accurate TST-based models for predicting battery performance in EVs, and (b) testing different TST architectures, and gaining insights into the difference of the predictive performance showcased by these different TST models. 

\section{Different TST Architectures}

\subsection{Working of Typical Encoder-Decoder Transformers}

In general, a \textit{vanilla}-transformer architecture for natural language processing (NLP) tasks consists of multiple encoder/decoder blocks with identical sub-layers and a globally pooled output layer \cite{vaswani2017attention,wen2022transformers}. The input data undergoes four transformations within an NLP-transformer. First, text inputs are preprocessed through truncation or padding and tokenization \cite{devlin2018bert}. Second, transformer networks use positional encoding (PE) to handle sequential data without recurrence. Third, a stack of encoder blocks employs multiheaded self-attention to calculate contextualized embeddings. Fourth, a stack of decoder blocks uses multiheaded cross-attention to generate human-like text by focusing on high-value cross-attention scores \cite{wen2022transformers}. This framework is known as the teacher-forcing method in training transformer models.

\begin{figure}[!ht]
	\centering
	\includegraphics[width=1\columnwidth]{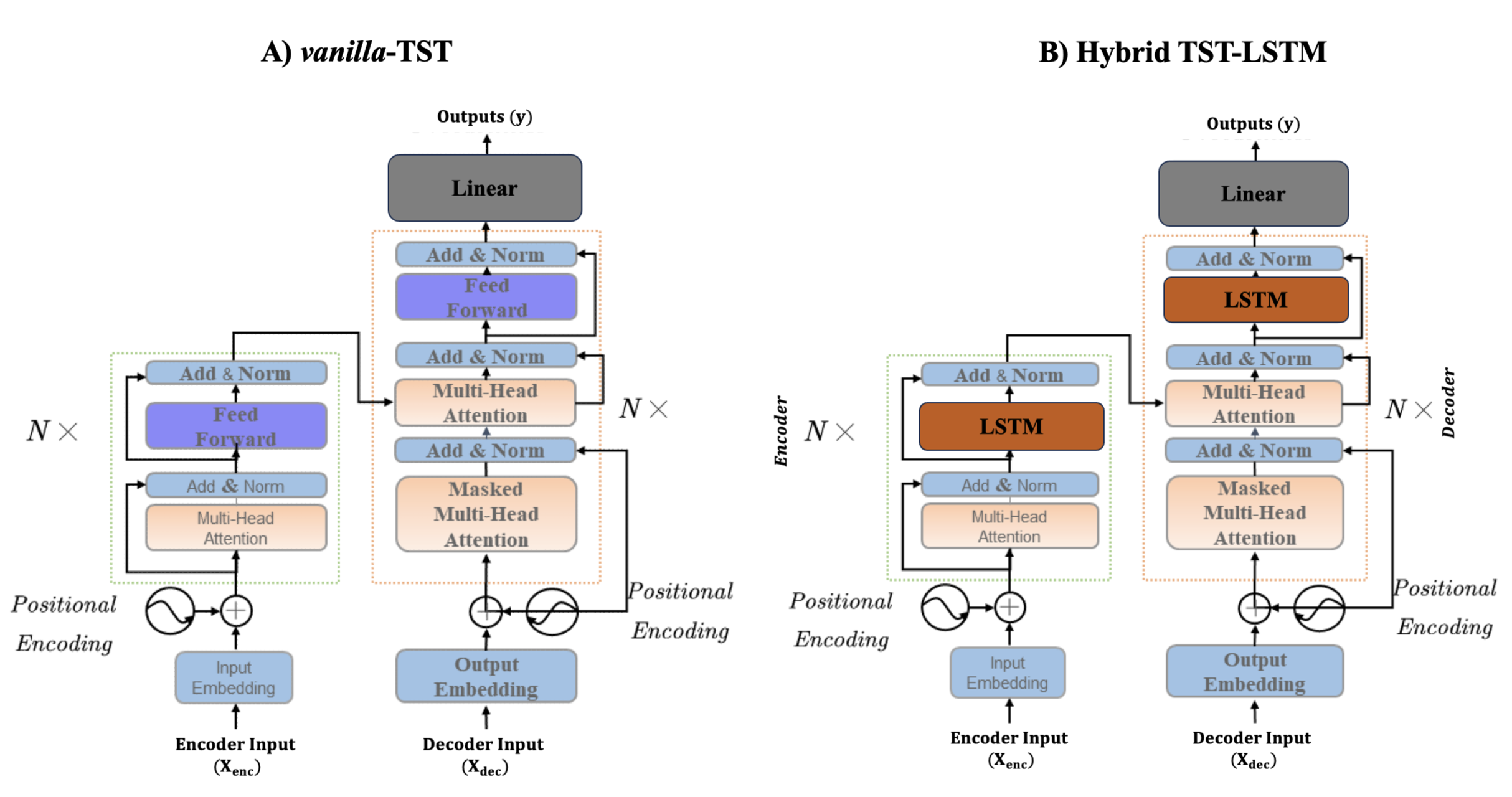}
	\caption{Schematic illustration of architecture for (a) v-TST, and (b) hybrid TST-LSTM models.}
	\label{vanilla_TST}
\end{figure}

Similar to NLP-transformers, TSTs utilize multiple encoder-decoder blocks, a multiheaded attention mechanism, and PE to process time-series dynamics comprising of state information. Specifically, TST takes in an input tensor (i.e., $[X_{t-W}, X_{t-W+1} ... X_{t}]$) and yields an output tensor with a prediction horizon of $H$ and dimension $v$ (i.e., $[y_{t+1}, y_{t+2}, ... y_{t+H}]$). After using PE, the input tensor is processed through a scaled-dot product query-key-value ([\textbf{Q},\textbf{K},\textbf{V}]) approach to compute attention scores as shown below \cite{vaswani2017attention,sitapure2023exploring}: 

\begin{equation}
	\begin{aligned}
		& A_{P,n} = \sum_{i}^{k} \lambda_{n,i} \textbf{V}&\\
		& \lambda_{n,i} = \frac{exp\left(\textbf{Q}^T\textbf{K}_i/\sqrt{D_k}\right)}{\sum_{j=1}^{k}exp\left(\textbf{Q}^T\textbf{K}_j/\sqrt{D_k}\right)} & \\ 
		& \sum_{i=1}^{k} \lambda_{n,i} =1& 
	\end{aligned}\label{QKV_model}
\end{equation}
where $A_{P,n}$ is the attention value for head $n$ in encoder block $P$, \textbf{Q}$\in \mathbb{R}^{D_k}$ are queries, \textbf{K}$\in \mathbb{R}^{D_k}$ stands for keys, and \textbf{V}$\in \mathbb{R}^{D_v}$ are values. Here, $D_k$ and $D_v$ are the dimensions of keys and values, respectively. Next, all the attention scores are processed by an FFN that captures the nonlinearity between inputs and outputs. This process is repeated in each encoder and decoder block. Previous literature works to provide a detailed explanation of the intricate working of TST \cite{wen2022transformers, zeng2022transformers, sitapure2023exploring}.

\subsection{Constructing Different TST Architectures}

\begin{figure}[!ht]
	\centering
	\includegraphics[width=0.65\columnwidth]{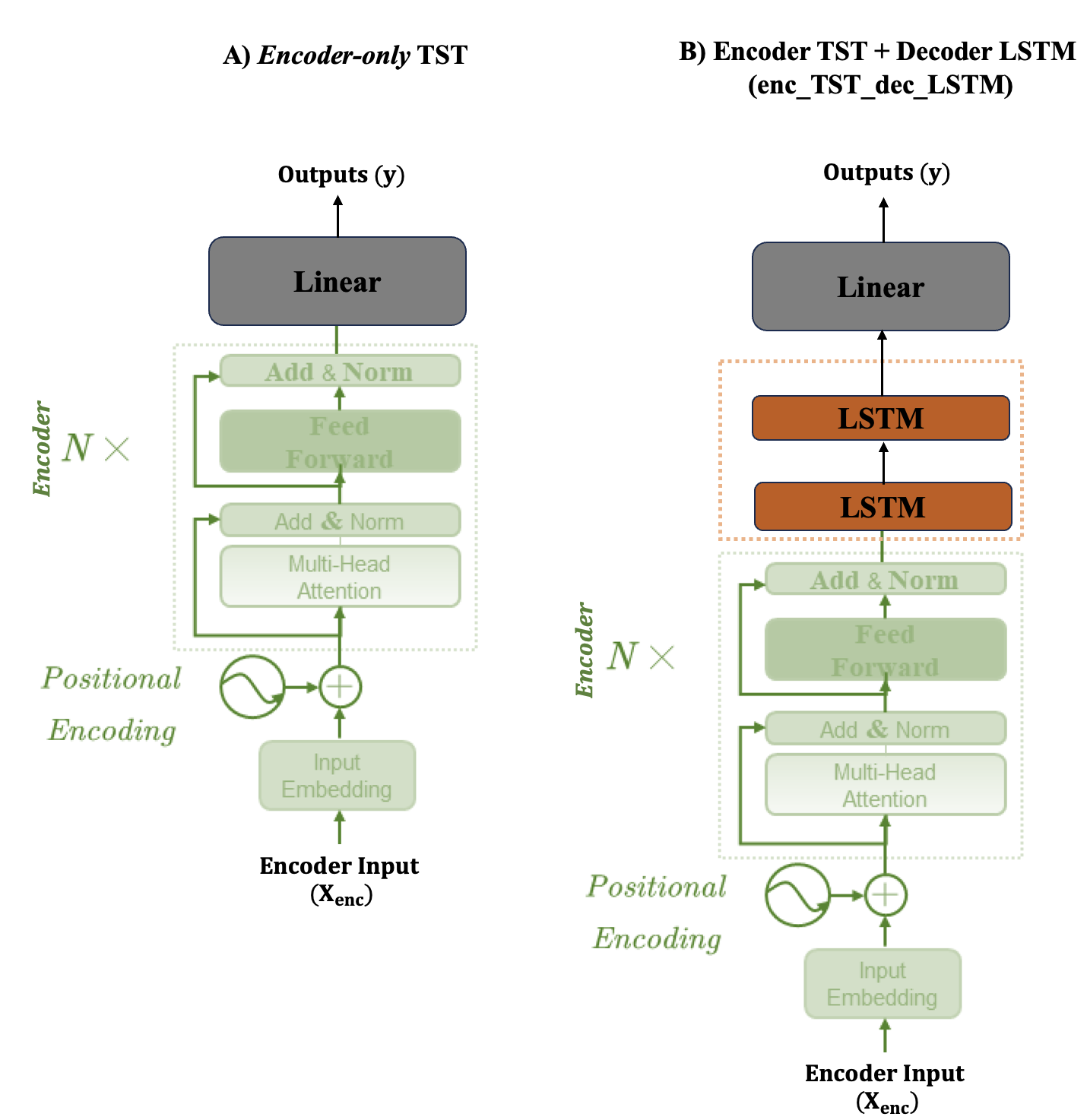}
	\caption{Schematic illustration of architecture for (a) enc-TST, and (b) enc-TST-dec-LSTM models.}
	\label{encoder_only_TST}
\end{figure}

As mentioned earlier, different TST architectures have been tested in this work. First, a straightforward encoder-only TST (enc-TST) is constructed as shown in Figure~\ref{encoder_only_TST}a. This model comprises only encoder blocks with the self-attention mechanism that are stacked on top of each other, which is followed by a final linear layer to predict the model outputs. Second, a \textit{vanilla}-TST (v-TST) is constructed as shown in Figure~\ref{vanilla_TST}a. In a v-TST, $N$ encoder blocks are stacked on each other which utilized a self-attention mechanism, and this is followed by $N$ decoder blocks that utilized a cross-attention mechanism. The entire encoder-decoder apparatus is connected to a final linear layer that predicts the system outputs. The above two models represent typical TST architectures that have gained attention in the past 2 years and have been demonstrated for a plethora of different applications \cite{sitapure2023exploring}. As explained before, these models leverage the combination of attention mechanism and positional encoding to provide high predictive performance \cite{sitapure2023crystalgpt}. 

Modifications to the above-mentioned traditional TST architectures can be envisioned that can potentially provide better predictive performance. For instance, LSTM networks that utilize explicit sequential-based model architecture are considered to be the gold standard for time-series predictions. Thus, if the LSTM network can be integrated with already impressive TST models, then potentially better predictive performance can be yielded from these new architectures. To this end, a hybrid TST-LSTM architecture is developed as a modification to the traditional v-TST as shown in Figure~\ref{vanilla_TST}b. Basically, the FFN in the v-TST comprises a linear network of neurons. However, in a hybrid TST-LSTM model, the FFN is replaced by LSTM layers. The rationale here is that the LSTM network, which utilizes an explicit sequential approach making it well-suited for time-series modeling, can augment the predictive capabilities of the v-TST. Another approach that follows the same rationale is developing an encoder-TST and decoder-LSTM (enc-TST-dec-LSTM) model. In this architecture, the entire set of decoder blocks is substituted with subsequently stacked LSTM layers as shown in Figure~\ref{encoder_only_TST}b. The rationale here is that the encoder blocks will utilize a self-attention mechanism to adeptly understand the interdependencies between system states, and then feed it to the LSTM-based decoder blocks that can track the temporal evolution of system states. Finally, considering the aforementioned TST architectures, four TST models were constructed, trained, and tested against a traditional LSTM model, which is considered the current SOTA model for time-series modeling. The architectural details for these models are mentioned in Table~\ref{table:models_comparison}. Although the model sizes (i.e., number of parameters) might vary between the models, the hyperparameters are kept as similar as possible (e.g., number of encoder-decoder blocks, number of attention heads, and others). Also, it is noteworthy to understand that having a large number of model parameters can be useful for capturing more subtle interdependencies between system states but also may require a large amount of training data to ensure appropriate estimation of all model parameters during model training. Thus, different TST models selected in this work also help us provide insight into the above aspect related to the number of model parameters. 

\begin{table}
\centering
\begin{tabular}{lccccc}
\toprule
 & enc-TST & enc-TST-dec-LSTM & v-TST & Hybrid TST-LSTM & LSTM \\
\midrule
\# Encoders & 4 & 4 & 4 & 4 & - \\
\# Decoders & 4 & 4 & 4 & 4 & - \\
\# Attention Heads & 8 & 8 & 8 & 8 & - \\
Inner Dimensions ($d_{model}$) & 128 & 128 & 128 & 128 & 128 \\
\# Neurons in FFN & 128 & 128 & 128 & 128 & - \\
\# Layers & - & - & - & - & 4 \\
\# Parameters & 400K & 950K & 1M & 3M & 475K \\
\bottomrule
\end{tabular}
\caption{Architectural details for different TST models and the reference LSTM model. }
\label{table:models_comparison}
\end{table}

\subsection{Data Preprocessing} 
As mentioned before, we utilized a dataset comprising 72 unique driving tests performed on a BMW i3 (60 Ah), which includes information about the environment, battery metrics, vehicle driving parameters, and heating circuit data every 0.1 seconds. Although the original dataset has information from 45+ different sensors, only relevant 15 features were selected. For example, the original dataset has sensor data for [Temperature Vent right, Temperature Vent central right, Temperature Vent central left, Temperature Vent right] that was averaged to create a new variable [Average Vent Temperature], which follows very similar time-series dynamics. For instance, all the vent sensors show a maximum deviation of 0.01\%, thereby suggesting that the inclusion of all the vent sensor data will not provide any new information to the ML models. This process was repeated for several different temperature sensors to reduce the number of redundant model features while retaining the information from the original features. 

\begin{figure}[!ht]
	\centering
	\includegraphics[width=0.85\columnwidth]{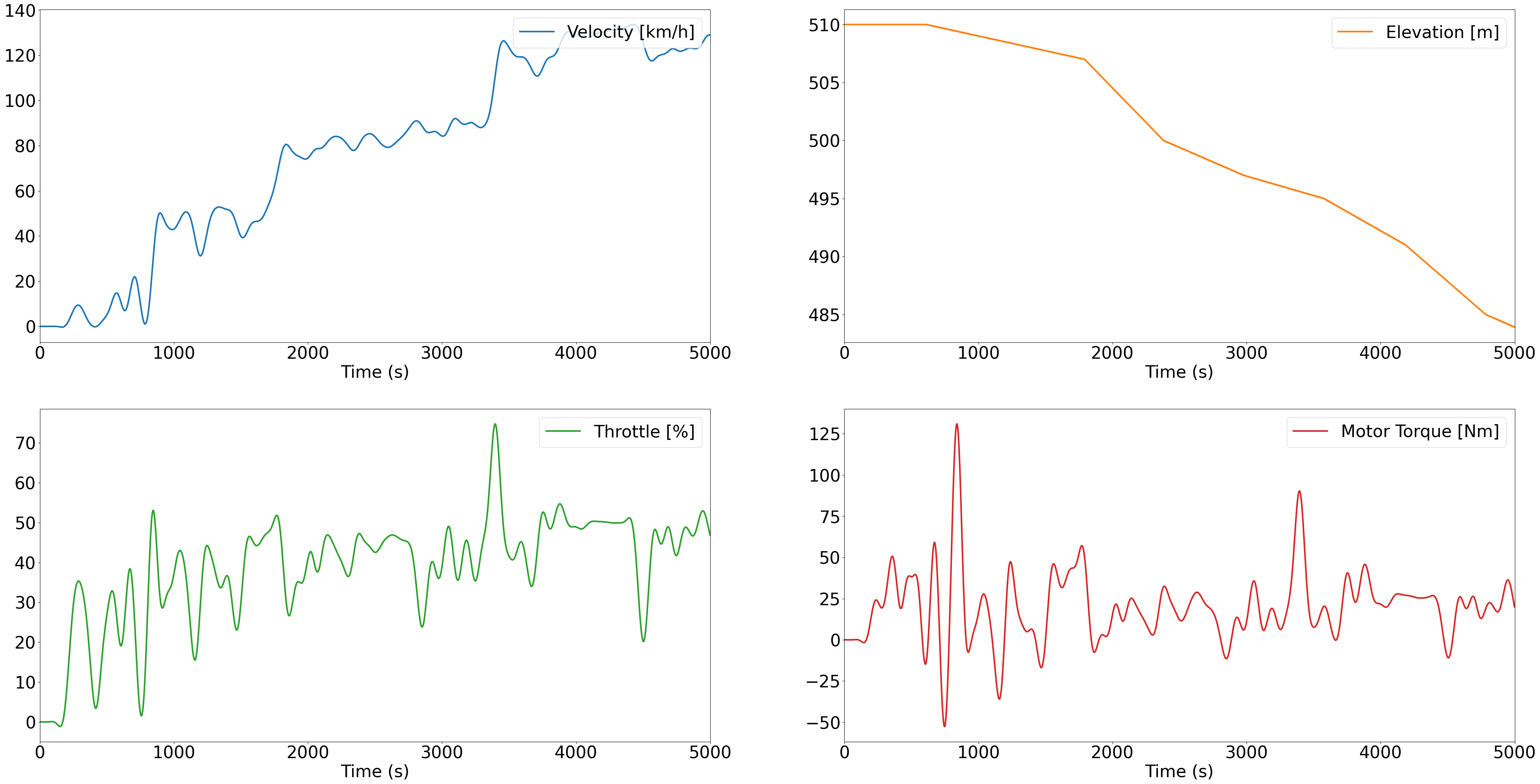}
	\caption{Temporal evolution of driving characteristics for an arbitrarily selected trip performed using the BMW i3.}
	\label{figure:driving_conditions}
\end{figure}

\begin{figure}[!ht]
	\centering
	\includegraphics[width=0.85\columnwidth]{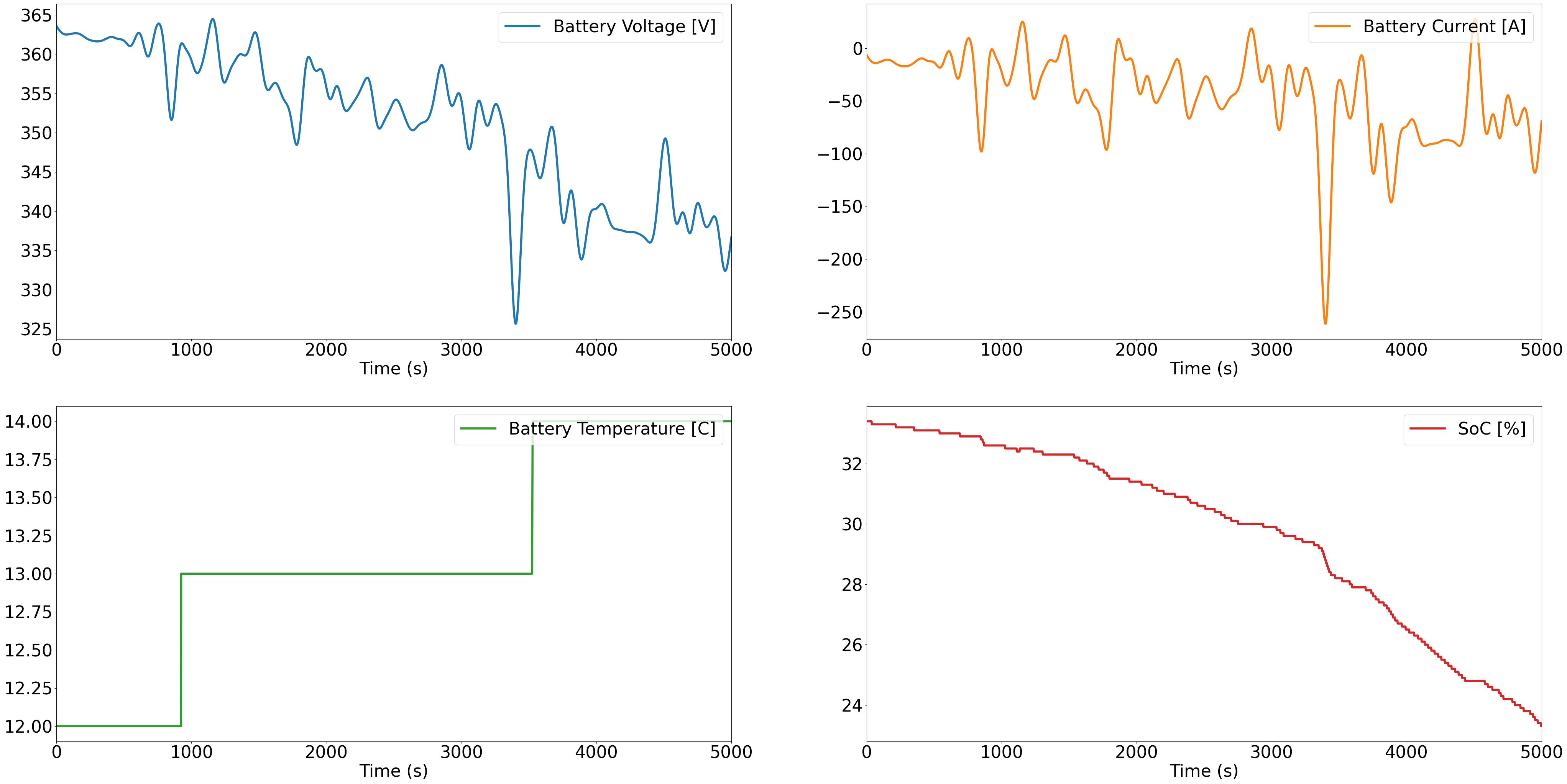}
	\caption{Temporal evolution of battery characteristics for an arbitrarily selected trip performed using the BMW i3.}
	\label{figure:battery_conditions}
\end{figure}

Furthermore, since all the sensor data is sampled at 0.1 seconds, there is considerable process noise, which can be detrimental to TST models. This is because TST models utilize attention mechanism to find interdependencies and trends in system data, and the inclusion of noisy data does not allow them to efficiently find patterns in the input data, which can lead to poor predictive performance. Thus, a Savitzky-Golay (Savgol) filter is utilized to denoise the sensor data for providing smoother inputs to the TST models without losing key process dynamics. Basically, The Savgol filter is a digital signal processing technique used to denoise process data by effectively removing noise while preserving important underlying trends and features \cite{gallagher2020savitzky}. It works by fitting a polynomial to local data points within a user-defined window, allowing engineers and analysts to identify and study the underlying patterns in the data \cite{rahman2016comparison}. The filter reduces random fluctuations, maintains data integrity, and handles irregular sampling, making it suitable for various process data denoising needs. Its tunable parameters and computational efficiency make it an attractive option for denoising large datasets commonly encountered in process monitoring and control, offering a valuable tool to analyze and interpret process behavior and trends. 

\begin{figure}[!ht]
	\centering
	\includegraphics[width=1\columnwidth]{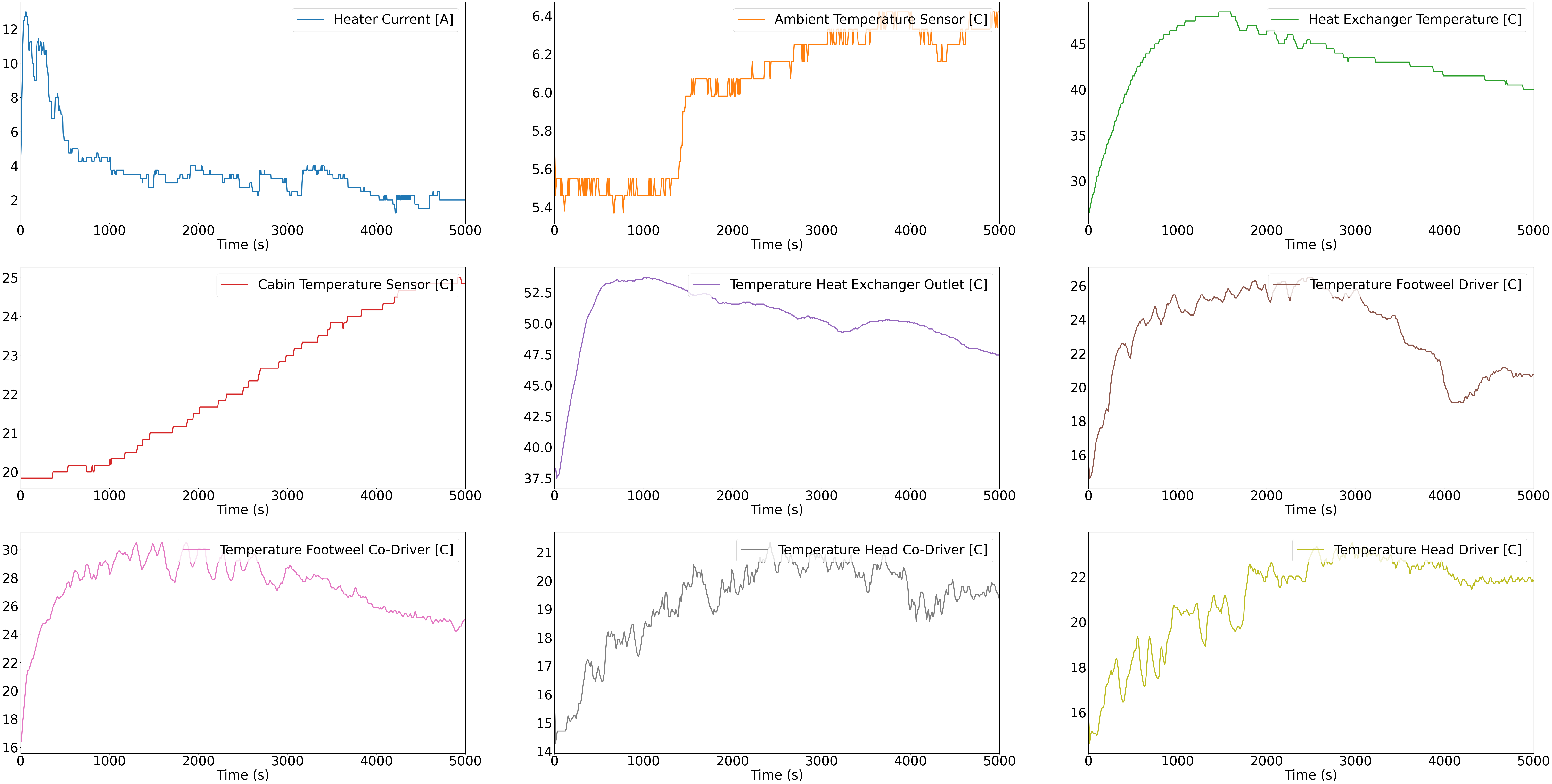}
	\caption{Temporal evolution of sensor data related to the heating circuit for an arbitrarily selected trip performed using the BMW i3.}
	\label{figure:heating_conditions}
\end{figure}

The resulting features can then be classified into three categories (i.e., driving, battery, and heating characteristics). For instance, Figures~\ref{figure:driving_conditions}-\ref{figure:heating_conditions} depict the temporal evolution of various system states for the above-mentioned three categories after the sensor data were denoised using a Savgol filter of order 2. Data for such 40+ trips were compiled into a large dataset with 4000 datapoints for training, 1000 datapoints for validation, and model testing. This dataset was shuffled to avoid any sequential bias and was stored as a \textit{PyTorch} dataset in a \textit{pickle} file. Further, each of the above models (Table~\ref{table:models_comparison}) was trained and tested using the aforementioned dataset to ensure a fair comparison of their predictive abilities. Basically, various sensor data were considered as model inputs for current and past $W$ time-steps, and the battery SOC and temperature were considered as outputs for $H$ time-steps in the future. 

\begin{table}[!ht]
\centering
\begin{tabular}{lcccccc}
\toprule
 & & \textbf{LSTM} & \textbf{enc-TST} & \textbf{\textbf{v-TST}} & \textbf{TST-LSTM} & \textbf{enc-TST-dec-LSTM} \\ 
 & \# Parameters & 475K & 400K & 1M & 3M & 950K \\
 \midrule
 \multirow{2}{*}{\centering } & Training & 20 & 32 & \textbf{11} & 18 & 51 \\
 & Validation & 21 & 33 & \textbf{17} & 19 & 68 \\
 $W=12$ and $H=6$ & Test & 22 & 34 & \textbf{17} & 24 & 72 \\
 & $R^2$ (Test) & 0.978 & 0.966 & \textbf{0.988} & 0.981 & 0.931 \\
\midrule
\multirow{2}{*}{\centering } & Training & 17 & 20 & \textbf{15} & 16 & 42 \\
 & Validation & 25 & 24 & \textbf{16} & 20 & 44 \\
$W=30$ and $H=6$ & Test & 26 & 33 & \textbf{18} & 22 & 53 \\
 & $R^2$ (Test) & 0.973 & 0.979 & \textbf{0.98} & 0.976 & 0.956 \\
\midrule
\multirow{2}{*}{\centering } & Training & 15 & 19 & \textbf{9} & 17 & 25 \\
 & Validation & 17 & 24 & \textbf{10} & 18 & 35 \\
$W=50$ and $H=30$ & Test & 22 & 26 & \textbf{11} & 23 & 39 \\
 & $R^2$ (Test) & 0.983 & 0.974 & \textbf{0.988} & 0.981 & 0.963 \\
\bottomrule
\end{tabular}
\caption{Compilation of training, validation, and testing results for different TST models and their comparison with LSTM model. The values in \textbf{bold} indicate the best-performing model for that case. }
\label{table:results_comparison}
\end{table}

\section{Results and Discussion}
\subsection{Comparison of Different TST Architectures}
Training, validation, and testing performance using the above dataset and the various TST architectures are tabulated in Table~\ref{table:results_comparison}. It can be seen that across all the different cases for different window sizes ($W$) and prediction horizons ($H$), the v-TST model outperforms all models consistently. Surprisingly, the traditional LSTM model comes in a close second and shows relatively good performance as compared to the rest of the TST models except v-TST. Meanwhile, the encoder-only TST model (enc-TST) shows adequate predictive performance. Interestingly, the rationale for developing hybrid TST-LSTM models (i.e., TST-LSTM, and enc-TST-dec-LSTM) does not show promise as indicated by the least accurate model (enc-TST-dec-LSTM) and an adequately good model (TST-LSTM). Moreover, the following general trend can be observed amongst the different TST models: 

\begin{equation}
    \text{v-TST} > \text{LSTM} > \text{TST-LSTM} > \text{enc-TST} > \text{enc-TST-dec-LSTM}
    \label{result:model_performance}
\end{equation}

It is worthwhile to suggest the possible reasons for the above trend in a sequential manner. Firstly, it is important to understand the reason v-TST shows great predictive capabilities. In a v-TST with $n$ attention heads, each attention head compute attention scores for each encoder and decoder block, which are then combined using the weights $c_j$ and processed through an FFN to approximate the input/output relationship. The depth and width of the v-TST enable it to capture complex representations of the input/output relationship \cite{sitapure2023crystalgpt}. Next, the parallelized nature of multiple attention heads in the v-TST allows it to break down a unified mapping function between inputs and outputs into multiple subspaces. Each attention head attends to different interdependencies among various system states when trained with input data for multiple systems. This process results in not only individual constituents but also shared subspace models between different system states. Basically, this distributed internal framework is essential for maintaining high accuracy when applying the v-TST model to a new prediction task. In this case, v-TST learns from sensor data for 30+ different driving trips by a BMW i3. When v-TST is tested for a new trip, it finds similarities between the current trip and previously learned trips and seamlessly interpolates to accurately predict the battery performance (i.e., SOC and battery temperature) \cite{chen2022transformer}. Secondly, the enc-TST shows relatively poor performance due to its limitations in capturing the full input-output relationship. Basically, enc-TST lacks the decoder component responsible for generating the output sequence, leading to the loss of information and reduced adaptability for sequence-to-sequence mapping. Thus, in enc-TST, only self-attention amongst the input tensor ($X_{enc}$) is utilized, and its connectivity to the output tensor ($X_{dec}$) is missing. In contrast, a v-TST allows bidirectional information flow and utilizes the `cross-attention mechanism' to focus on relevant parts of the input sequence ($X_{enc}$) and connect it with the output sequence ($X_{dec}$) during model training. A detailed explanation of the working of TSTs is available in the literature \cite{sitapure2023crystalgpt, sitapure2023exploring, lim2021temporal}

Thirdly, it is interesting to ponder on the reasons for the abysmal performance of the enc-TST-dec-LSTM model. The lack of predictive capabilities can be attributed to some key factors. For instance, the hybrid architecture of the enc-TST-dec-LSTM model introduces additional complexity and mismatch in the symmetry of the architecture, which can hinder the model's ability to generalize well to new data. Further, the combination of encoder-TST and decoder-LSTM may not facilitate smooth information flow throughout the model. More importantly, there is no `cross-attention' module in enc-TST-dec-LSTM which disables the model from learning the continuity between input ($X_{enc}$) and output ($X_{dec}$) tensors during model training, which is an essential aspect of training transformer models. 
That being said, the reason for the adequate performance of the TST-LSTM model, wherein the FFN in a v-TST is replaced by LSTM layers, is a bit more interesting. In addition to the architectural mismatch, there can be three key reasons for an inferior performance. Basically, the attention mechanism in TSTs is a powerful feature that enables the model to focus on relevant parts of the input sequence while generating the output sequence. Replacing the FFN with LSTM layers might lead to the loss of this crucial attention mechanism, affecting the model's ability to handle complex sequences effectively. For instance, the encoder/decoder part of TST utilizes PE to time-stamp the features in the input tensor using sinusoidal transformations, which are then processed via the attention mechanism. However, the LSTM is adept at handling time series in an explicit manner, which is lost during the PE process, thereby negating the benefit of incorporating an LSTM layer instead of FFN. Also, even though there is a `cross-attention' between encoder and decoder blocks (similar to v-TST), the discontinuity in the attention mechanism during the FFN module may lead to poor training. Next, replacing the FFN with LSTM layers introduces a higher number of parameters for the same amount of training data as evident in Table~\ref{table:models_comparison}. This increased parameter count can lead to overfitting, especially if the dataset is limited, and the model might struggle to generalize well on unseen data. Furthermore, the hybrid model nature of TST-LSTM architecture might be more challenging to train and tune due to the combination of two different architectures. 

Lastly, we explore why LSTM can perform better than enc-TST and other hybrid TST-LSTM models. LSTMs are designed specifically for sequential data, and they have a built-in ability to process sequences in a step-by-step manner, retaining information from previous time steps through hidden states. LSTMs utilize recurrent connections that allow them to remember past information and update their hidden states accordingly \cite{sitapure2023require}. Also, LSTMs recurrent connections help in leveraging the limited data more effectively for learning patterns and trends. In simpler terms, the LSTM network has a lesser number of model parameters for the same amount of training data.  Furthermore, the symmetry of a standalone LSTM network as compared to the hybrid TST-LSTM networks might allow for more effective training. That being said, these characteristics are not enough for the LSTM model to supersede the performance of v-TST. As mentioned before, v-TST can break down a unified mapping function between inputs and outputs into more adaptable subspaces that can be dynamically weighted to make better time-series predictions. On the contrary, LSTM networks attempt to find a single unified function between inputs and outputs, thereby being prone to slight overfitting and poor generalizability \cite{sitapure2023crystalgpt}. 

\subsection{Effect of Window Size ($W$)}

The other aspect in Table~\ref{table:results_comparison} is the performance difference for changing $W$. It is evident that model performance for all the TST models and LSTM model increases for larger $W$. This is not a surprising observation and has been shown to be true across previous TST models \cite{dai2019transformer}. Essentially, larger $W$ allows the model to capture longer contextual dependencies, enabling a better understanding of the broader relationships between distant tokens. This enhanced modeling of long sequences reduces positional bias and improves the attention mechanism's ability to capture long-range dependencies. The increased capacity provided by larger windows allows for more comprehensive feature representation, leading to improved performance. Moreover, a larger window size also enables more continuity between the input and output tensors (i.e., $X_{enc}$ and $X_{dec}$) during model training, thereby allowing the TST model to better understand the interdependencies between system states.

\begin{figure}[!ht]
	\centering
	\includegraphics[width=0.85\columnwidth]{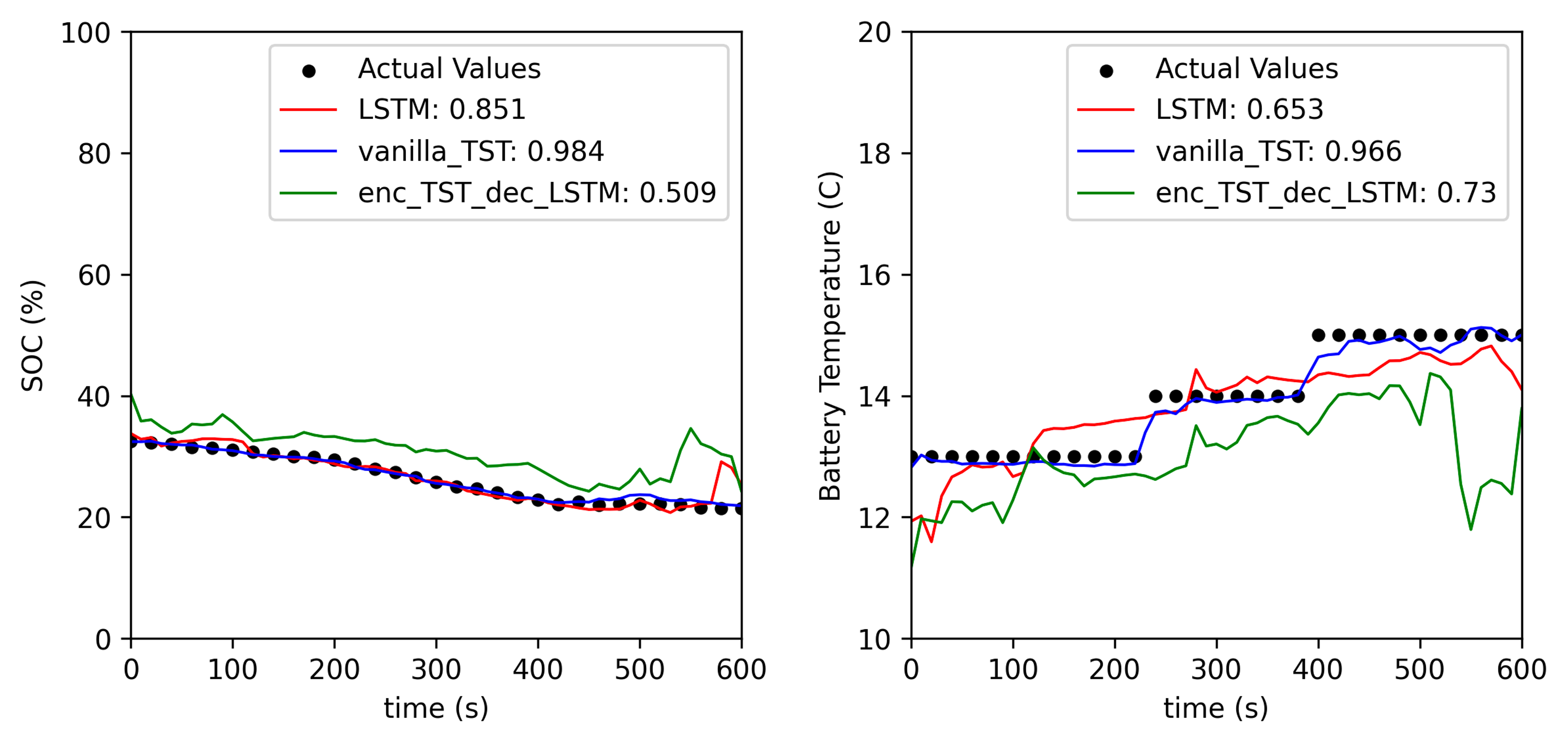}
	\caption{Temporal evolution of key battery metrics (i.e., SOC and temperature) as predicted by different TST models for $W=12$.}
	\label{figure:window_12}
\end{figure}

\begin{figure}[!ht]
	\centering
	\includegraphics[width=0.85\columnwidth]{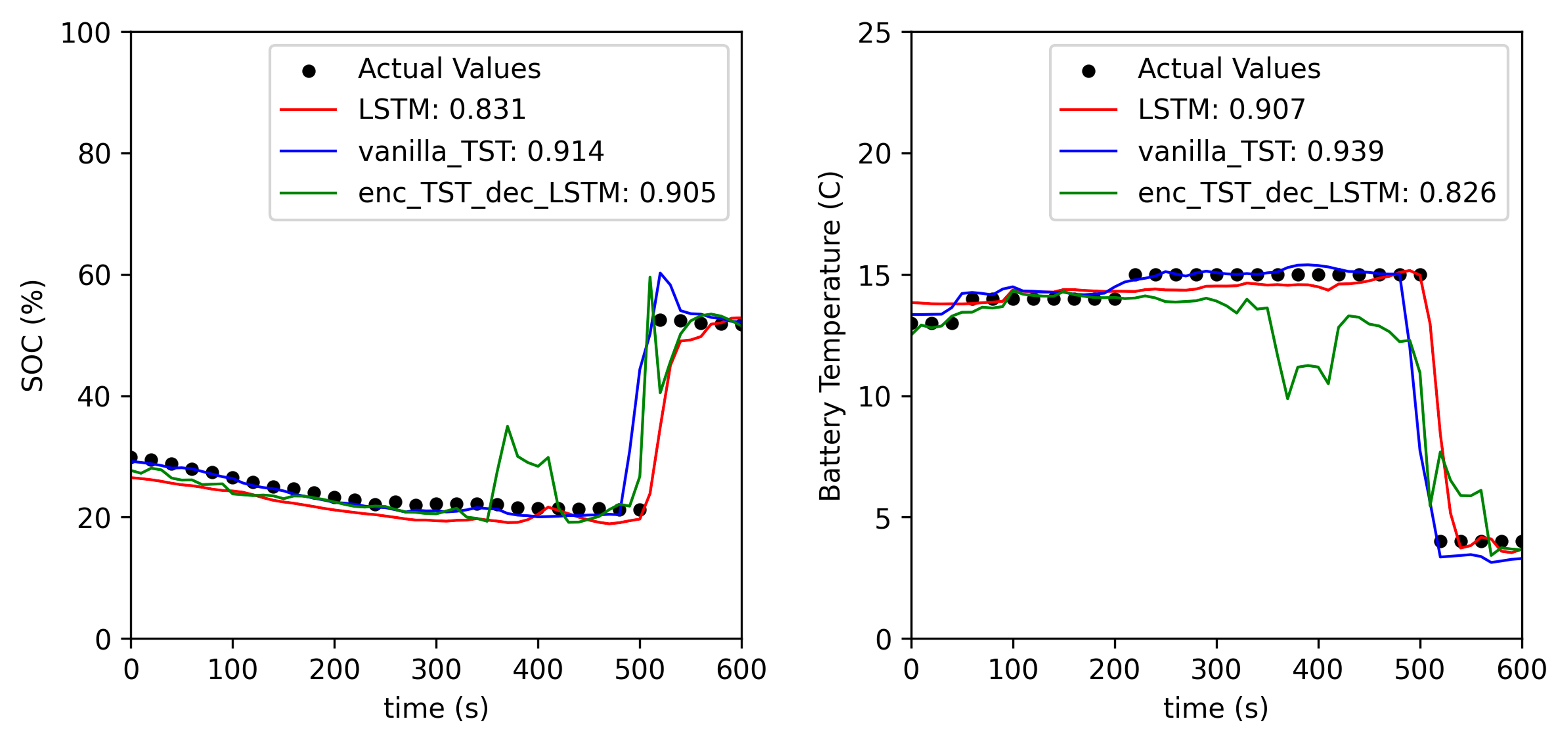}
	\caption{Temporal evolution of key battery metrics (i.e., SOC and temperature) as predicted by different TST models for $W=30$}
	\label{figure:window_30}
\end{figure}

This is evident from Figures~\ref{figure:window_12} and \ref{figure:window_30}, which showcase the model predictions for key battery parameters (i.e., SOC and temperature) for an arbitrarily selected trip performed in BMW i3. First, it is evident that the v-TST model shows the highest $R^2$ values among all the models for both cases. Also, it can be seen that the temporal predictions by enc-TST-dec-LSTM models have a large oscillating and lack accuracy as indicated in Table~\ref{table:results_comparison}. Further, these oscillations can be attributed to the highly asymmetrical and complex architecture of enc-TST-dec-LSTM as explained in the previous section. Lastly, the LSTM model performs adequately well as seen in Table~\ref{table:results_comparison}, and although it shows some oscillating behavior for $W=12$, the oscillations are minimized for $W=30$. Again, this occurs due to the fact that for higher $W$, all the models have access to longer input sequences allowing them to better gauge the trajectory of different system states. 

\begin{rmk}
    In Figures~\ref{figure:window_12} and \ref{figure:window_30}, only three models are depicted to avoid overcrowding of results. Moreover, v-TST is picked as it shows the least testing error in Table~\ref{table:results_comparison}. While enc-TST-dec-LSTM is chosen as it shows the highest testing errors. And the LSTM model is chosen as it is a baseline model that is the current SOTA for time-series predictions. 
\end{rmk}

\subsection{Discussion}

The above sections demonstrated the capability of different TST models for achieving high predictive performance in complex time-series modeling tasks. Building on the above models and some of the ones developed in the literature \cite{sitapure2023crystalgpt, sitapure2023exploring}, TST-based time-series modeling can be applied to a plethora of different applications. For example, another approach that can hugely benefit from the incorporation of various TST models is the hybrid modeling of various chemical, mechanical, and electrical systems. Basically, Hybrid models that combine system-agnostic first principles with system-specific data-driven parameters have been demonstrated for fermentation, fracking, and other chemical processes  \cite{bangi2020deep, bangi2022physics, shah2022deep, lee2020development}. For instance, in industrial fermentation, physics-based reactor models inclusive of mass and heat transfer can be integrated with a system-specific deep neural network (DNN) that predicts the fermentation kinetics to simulate the process. That said, DNNs can exhibit limitations in accurately predicting complex time-varying parameters commonly found in intricate chemical processes. Moreover, DNNs lack the capability to leverage contextual information regarding the trajectory of process dynamics. Empowered by multiheaded attention mechanisms and PE, TST models can potentially perform better for hybrid modeling of intricate chemical and mechanical engineering applications. 

Another modeling paradigm wherein attention-based TST models can be utilized is in multiscale models, which integrate physical laws at different time and length scales to uncover mechanisms governing observable phenomena. For instance, a considerable number of multiscale models have been developed in the past for particulate systems, thin-film depositions, pulping processes, biochemical processes, and others. \cite{SITAPURE2020127905, sitapure2021cfd, sitapure2022neural, kwon2013modeling_aggregate,kwon2014crystal,kwon2014enhancing, choi2019modeling, choi2020multiscale}. In these cases, there are a number of process parameters (e.g., kinetic constants, growth rates, product characteristics, etc.) that connect the different time and length scales, and it can often be non-trivial to incorporate all these interactions in a high-fidelity multiscale model. In such cases, TST models can utilize an attention mechanism to emphasize the most dominating interacting parameters to reduce model complexity. On the other hand, similar attention-based TST models can be utilized in conjunction with a multiscale model to reduce the plant-model mismatch in some of these models. For example, an existing multiscale model might consider
1-2 interacting parameters (out of 10), thereby resulting in certain deviations from experimental/pilot/process data. Thus, this multiscale model can be coupled with a TST that utilizes available process data to correct these deviations leading to more accurate yet computationally non-prohibitive multiscale models. 

\section{Conclusions}

While significant efforts have been devoted to developing new battery materials and chemistries, accurately predicting crucial battery parameters (e.g., SOC and temperature), remains a challenge. Thus to explore SOTA data-driven approaches for accurate prediction of battery characteristics, in this, work, various TST architectures have been explored. Specifically, four different TST models (i.e., enc-TST, v-TST, TST-LSTM, and enc-TST-dec-LSTM) were trained and tested against a traditional LSTM model. A practical dataset comprising 72 distinct driving trips performed in a BMW i3 (60 Ah), which records sensor data for environmental conditions, driving, and battery characteristics is utilized to predict SOC and battery temperature for $H$ future time-steps. The testing results show an interesting trend in model performance (i.e., \text{v-TST} > \text{LSTM} > \text{TST-LSTM} > \text{enc-TST} > \text{enc-TST-dec-LSTM}). This behavior is attributed to the fact that hybrid TST architectures like TST-LSTM and enc-TST-dec-LSTM result in asymmetrical models that also lose critical information from attention scores due to the incorporation of distinctly different LSTM layers. That being said, all the models show better performance (i.e., lower errors) for increasing $W$, as longer input sequences provide more contextual information leading to better model training. Overall, the present work showcases an interesting phenomenon, wherein more sophisticated and complex hybrid TST-LSTM architectures underperform as compared to their traditional TST and LSTM counterparts. Further, the above models demonstrate the utilization of TST models to enhance the accuracy of battery characteristics predictions, contributing to the advancement of battery management systems in electric vehicles.

\newpage

\bibliographystyle{unsrt}  
\bibliography{BMW_paper}  

\begin{thebibliography}{10}

\bibitem{wang_review_2018}
Aiping Wang, Sanket Kadam, Hong Li, Siqi Shi, and Yue Qi.
\newblock Review on modeling of the anode solid electrolyte interphase ({SEI})
  for lithium-ion batteries.
\newblock {\em npj Computational Materials}, 4(1):1--26, March 2018.
\newblock Number: 1 Publisher: Nature Publishing Group.

\bibitem{nitta2015li}
Naoki Nitta, Feixiang Wu, Jung~Tae Lee, and Gleb Yushin.
\newblock Li-ion battery materials: present and future.
\newblock {\em Materials Today}, 18(5):252--264, 2015.

\bibitem{nykvist2015rapidly}
Bj{\"o}rn Nykvist and M{\aa}ns Nilsson.
\newblock Rapidly falling costs of battery packs for electric vehicles.
\newblock {\em Nature Climate Change}, 5(4):329--332, 2015.

\bibitem{parekh2022critical}
Mihit~H Parekh, Suyash Oka, Jodie Lutkenhaus, and Vilas~G Pol.
\newblock Critical-point-dried, porous, and safer aramid nanofiber separator
  for high-performance durable lithium-ion batteries.
\newblock {\em ACS Applied Materials \& Interfaces}, 14(25):29176--29187, 2022.

\bibitem{manthiram2020reflection}
Arumugam Manthiram.
\newblock A reflection on lithium-ion battery cathode chemistry.
\newblock {\em Nature communications}, 11(1):1550, 2020.

\bibitem{zhao2021review}
Yanyan Zhao, Oliver Pohl, Anand~I Bhatt, Gavin~E Collis, Peter~J Mahon, Thomas
  R{\"u}ther, and Anthony~F Hollenkamp.
\newblock A review on battery market trends, second-life reuse, and recycling.
\newblock {\em Sustainable Chemistry}, 2(1):167--205, 2021.

\bibitem{sitapure2020computational}
Niranjan Sitapure, Hyeonggeon Lee, Francisco Ospina-Acevedo, Perla~B Balbuena,
  Sungwon Hwang, and Joseph~S Kwon.
\newblock A computational approach to characterize formation of a passivation
  layer in lithium metal anodes.
\newblock {\em AIChE Journal}, e17073, 2020.

\bibitem{lee2021multiscale}
Hyeonggeon Lee, Niranjan Sitapure, Sungwon Hwang, and Joseph Sang-Il Kwon.
\newblock Multiscale modeling of dendrite formation in lithium-ion batteries.
\newblock {\em Computers \& Chemical Engineering}, page 107415, 2021.

\bibitem{pozzi2020optimal}
Andrea Pozzi, Marcello Torchio, Richard~D Braatz, and Davide~M Raimondo.
\newblock Optimal charging of an electric vehicle battery pack: A real-time
  sensitivity-based model predictive control approach.
\newblock {\em Journal of Power Sources}, 461:228133, 2020.

\bibitem{hwang2022model}
Gyuyeong Hwang, Niranjan Sitapure, Jiyoung Moon, Hyeonggeon Lee, Sungwon Hwang,
  and Joseph~Sang Kwon.
\newblock Model predictive control of lithium-ion batteries: Development of
  optimal charging profile for reduced intracycle capacity fade using an
  enhanced single particle model ({SPM}) with first-principled
  chemical/mechanical degradation mechanisms.
\newblock {\em Chemical Engineering Journal}, 435:134768, 2022.

\bibitem{torchio_real-time_2015}
Marcello Torchio, Nicolas~A. Wolff, Davide~M. Raimondo, Lalo Magni, Ulrike
  Krewer, R.~Bushan Gopaluni, Joel~A. Paulson, and Richard~D. Braatz.
\newblock Real-time model predictive control for the optimal charging of a
  lithium-ion battery.
\newblock In {\em 2015 {American} {Control} {Conference} ({ACC})}, pages
  4536--4541, Chicago, IL, USA, July 2015. IEEE.

\bibitem{suthar_optimal_2013}
Bharatkumar Suthar, Venkatasailanathan Ramadesigan, Paul W.~C. Northrop,
  Bhushan Gopaluni, Shriram Santhanagopalan, Richard~D. Braatz, and Venkat~R.
  Subramanian.
\newblock Optimal control and state estimation of lithium-ion batteries using
  reformulated models.
\newblock In {\em 2013 {American} {Control} {Conference}}, pages 5350--5355,
  June 2013.
\newblock ISSN: 2378-5861.

\bibitem{dehghani2019study}
AR~Dehghani-Sanij, E~Tharumalingam, MB~Dusseault, and R~Fraser.
\newblock Study of energy storage systems and environmental challenges of
  batteries.
\newblock {\em Renewable and Sustainable Energy Reviews}, 104:192--208, 2019.

\bibitem{keil2015aging}
Peter Keil and Andreas Jossen.
\newblock Aging of lithium-ion batteries in electric vehicles: Impact of
  regenerative braking.
\newblock {\em World Electric Vehicle Journal}, 7(1):41--51, 2015.

\bibitem{steinstraeter2020range}
Matthias Steinstraeter, Marcel Lewke, Johannes Buberger, Tobias Hentrich, and
  Markus Lienkamp.
\newblock Range extension via electrothermal recuperation.
\newblock {\em World Electric Vehicle Journal}, 11(2):41, 2020.

\bibitem{tian2017understanding}
Zhiyong Tian, Lai Tu, Chen Tian, Yi~Wang, and Fan Zhang.
\newblock Understanding battery degradation phenomenon in real-life electric
  vehicle use based on big data.
\newblock In {\em 2017 3rd International Conference on Big Data Computing and
  Communications (BIGCOM)}, pages 334--339. IEEE, 2017.

\bibitem{lv2022machine}
Chade Lv, Xin Zhou, Lixiang Zhong, Chunshuang Yan, Madhavi Srinivasan, Zhi~Wei
  Seh, Chuntai Liu, Hongge Pan, Shuzhou Li, and Yonggang Wen.
\newblock Machine learning: an advanced platform for materials development and
  state prediction in lithium-ion batteries.
\newblock {\em Advanced Materials}, 34(25):2101474, 2022.

\bibitem{severson2019data}
Kristen~A Severson, Peter~M Attia, Norman Jin, Nicholas Perkins, Benben Jiang,
  Zi~Yang, Michael~H Chen, Muratahan Aykol, Patrick~K Herring, Dimitrios
  Fraggedakis, et~al.
\newblock Data-driven prediction of battery cycle life before capacity
  degradation.
\newblock {\em Nature Energy}, 4(5):383--391, 2019.

\bibitem{ng2020predicting}
Man-Fai Ng, Jin Zhao, Qingyu Yan, Gareth~J Conduit, and Zhi~Wei Seh.
\newblock Predicting the state of charge and health of batteries using
  data-driven machine learning.
\newblock {\em Nature Machine Intelligence}, 2(3):161--170, 2020.

\bibitem{bhadriraju2019machine}
Bhavana Bhadriraju, Abhinav Narasingam, and Joseph~S Kwon.
\newblock Machine learning-based adaptive model identification of systems:
  Application to a chemical process.
\newblock {\em Chemical Engineering Research and Design}, 152:372--383, 2019.

\bibitem{Bhadriraju2019}
Bhavana Bhadriraju, Abhinav Narasingam, and Joseph~Sang Kwon.
\newblock Machine learning-based adaptive model identification of systems:
  application to a chemical process.
\newblock {\em Chemical Engineering Research and Design}, 152:372--383, 2019.

\bibitem{bhadriraju2021oasis}
Bhavana Bhadriraju, Joseph~Sang Kwon, and Faisal Khan.
\newblock {OASIS}-{P}: Operable adaptive sparse identification of systems for
  fault prognosis of chemical processes.
\newblock {\em Journal of Process Control}, 107:114--126, 2021.

\bibitem{bhadriraju2023adaptive}
Bhavana Bhadriraju, Joseph Sang-Il Kwon, and Faisal Khan.
\newblock An adaptive data-driven approach for two-timescale dynamics
  prediction and remaining useful life estimation of li-ion batteries.
\newblock {\em Computers \& Chemical Engineering}, page 108275, 2023.

\bibitem{park2020lstm}
Kyungnam Park, Yohwan Choi, Won~Jae Choi, Hee-Yeon Ryu, and Hongseok Kim.
\newblock Lstm-based battery remaining useful life prediction with
  multi-channel charging profiles.
\newblock {\em IEEE Access}, 8:20786--20798, 2020.

\bibitem{GPT4_technical_report}
{OpenAI}.
\newblock {GPT}-4 technical report.
\newblock 2303.08774, 2023.

\bibitem{devlin2018bert}
Jacob Devlin, Ming-Wei Chang, Kenton Lee, and Kristina Toutanova.
\newblock {BERT}: Pre-training of deep bidirectional transformers for language
  understanding.
\newblock {\em arXiv preprint}, 1810.04805, 2018.

\bibitem{vaswani2017attention}
Ashish Vaswani, Noam Shazeer, Niki Parmar, Jakob Uszkoreit, Llion Jones,
  Aidan~N Gomez, {\L}ukasz Kaiser, and Illia Polosukhin.
\newblock Attention is all you need.
\newblock {\em Advances in Neural Information Processing Systems}, 30, 2017.

\bibitem{radford2019language}
Alec Radford, Jeffrey Wu, Rewon Child, David Luan, Dario Amodei, and Ilya
  Sutskever.
\newblock Language models are unsupervised multitask learners.
\newblock {\em OpenAI blog}, 1(8):9, 2019.

\bibitem{brown2020language}
Tom Brown, Benjamin Mann, Nick Ryder, Melanie Subbiah, Jared~D Kaplan, Prafulla
  Dhariwal, Arvind Neelakantan, Pranav Shyam, Girish Sastry, Amanda Askell,
  et~al.
\newblock Language models are few-shot learners.
\newblock {\em Advances in Neural Information Processing Systems},
  33:1877--1901, 2020.

\bibitem{shoeybi2019megatron}
Mohammad Shoeybi, Mostofa Patwary, Raul Puri, Patrick LeGresley, Jared Casper,
  and Bryan Catanzaro.
\newblock Megatron-{LM}: Training multi-billion parameter language models using
  model parallelism.
\newblock {\em arXiv preprint}, 1909.08053, 2019.

\bibitem{sitapure2023exploring}
Niranjan Sitapure and Joseph Sang-Il Kwon.
\newblock Exploring the potential of time-series transformers for process
  modeling and control in chemical systems: an inevitable paradigm shift?
\newblock {\em Chemical Engineering Research and Design}, 194:461--477, 2023.

\bibitem{wen2022transformers}
Qingsong Wen, Tian Zhou, Chaoli Zhang, Weiqi Chen, Ziqing Ma, Junchi Yan, and
  Liang Sun.
\newblock Transformers in time series: A survey.
\newblock {\em arXiv preprint}, 2202.07125, 2022.

\bibitem{sitapure2023crystalgpt}
Niranjan Sitapure and Joseph~Sang Kwon.
\newblock Crystal{GPT}: Enhancing system-to-system transferability in
  crystallization prediction and control using time-series-transformers.
\newblock {\em Computers \& Chemical Engineering}, 177:108339, 2023.

\bibitem{lim2021temporal}
Bryan Lim, Sercan~{\"O} Ar{\i}k, Nicolas Loeff, and Tomas Pfister.
\newblock Temporal fusion transformers for interpretable multi-horizon time
  series forecasting.
\newblock {\em International Journal of Forecasting}, 37(4):1748--1764, 2021.

\bibitem{bmw_dataset}
Matthias Steinstraeter, Johannes Buberger, and Dimitar Trifonov.
\newblock Battery and heating data in real driving cycles.
\newblock 2020.

\bibitem{zeng2022transformers}
Ailing Zeng, Muxi Chen, Lei Zhang, and Qiang Xu.
\newblock Are transformers effective for time series forecasting?
\newblock {\em arXiv preprint}, 2205:13504, 2022.

\bibitem{gallagher2020savitzky}
Neal~B Gallagher.
\newblock Savitzky-golay smoothing and differentiation filter.
\newblock {\em Eigenvector Research Incorporated}, 2020.

\bibitem{rahman2016comparison}
Md~Shahinoor Rahman, Liping Di, Ranjay Shrestha, G~Yu Eugene, Li~Lin, Lingjun
  Kang, and Meixia Deng.
\newblock Comparison of selected noise reduction techniques for modis daily
  ndvi: An empirical analysis on corn and soybean.
\newblock In {\em 2016 Fifth International Conference on Agro-Geoinformatics
  (Agro-Geoinformatics)}, pages 1--5. IEEE, 2016.

\bibitem{chen2022transformer}
Daoquan Chen, Weicong Hong, and Xiuze Zhou.
\newblock Transformer network for remaining useful life prediction of
  lithium-ion batteries.
\newblock {\em IEEE Access}, 10:19621--19628, 2022.

\bibitem{sitapure2023require}
Niranjan Sitapure and Joseph~S Kwon.
\newblock Require process control? {LSTM}c is all you need!
\newblock {\em arXiv preprint}, 2306:07510, 2023.

\bibitem{dai2019transformer}
Zihang Dai, Zhilin Yang, Yiming Yang, Jaime Carbonell, Quoc~V Le, and Ruslan
  Salakhutdinov.
\newblock Transformer-{XL}: Attentive language models beyond a fixed-length
  context.
\newblock {\em arXiv preprint}, 1901.02860, 2019.

\bibitem{bangi2020deep}
Mohammed Saad~Faizan Bangi and Joseph~S Kwon.
\newblock Deep hybrid modeling of chemical process: application to hydraulic
  fracturing.
\newblock {\em Computers \& Chemical Engineering}, 134:106696, 2020.

\bibitem{bangi2022physics}
Mohammed Saad~Faizan Bangi, Katy Kao, and Joseph~Sang Kwon.
\newblock Physics-informed neural networks for hybrid modeling of lab-scale
  batch fermentation for $\beta$-carotene production using saccharomyces
  cerevisiae.
\newblock {\em Chemical Engineering Research and Design}, 179:415--423, 2022.

\bibitem{shah2022deep}
Parth Shah, M~Ziyan Sheriff, Mohammed Saad~Faizan Bangi, Costas Kravaris,
  Joseph~Sang Kwon, Chiranjivi Botre, and Junichi Hirota.
\newblock Deep neural network-based hybrid modeling and experimental validation
  for an industry-scale fermentation process: Identification of time-varying
  dependencies among parameters.
\newblock {\em Chemical Engineering Journal}, 441:135643, 2022.

\bibitem{lee2020development}
Dongheon Lee, Arul Jayaraman, and Joseph~S Kwon.
\newblock Development of a hybrid model for a partially known intracellular
  signaling pathway through correction term estimation and neural network
  modeling.
\newblock {\em PLoS Computational Biology}, 16(12):e1008472, 2020.

\bibitem{SITAPURE2020127905}
Niranjan Sitapure, Robert Epps, Milad Abolhasani, and Joseph~S Kwon.
\newblock Multiscale modeling and optimal operation of millifluidic synthesis
  of perovskite quantum dots: towards size-controlled continuous manufacturing.
\newblock {\em Chemical Engineering Journal}, page 127905, 2020.

\bibitem{sitapure2021cfd}
Niranjan Sitapure, Robert~W Epps, Milad Abolhasani, and Joseph~S Kwon.
\newblock {CFD}-based computational studies of quantum dot size control in slug
  flow crystallizers: Handling slug-to-slug variation.
\newblock {\em Industrial \& Engineering Chemistry Research},
  60(13):4930--4941, 2021.

\bibitem{sitapure2022neural}
Niranjan Sitapure and Joseph Sang-Il Kwon.
\newblock Neural network-based model predictive control for thin-film chemical
  deposition of quantum dots using data from a multiscale simulation.
\newblock {\em Chemical Engineering Research and Design}, 183:595--607, 2022.

\bibitem{kwon2013modeling_aggregate}
Joseph~S Kwon, Michael Nayhouse, Panagiotis~D Christofides, and Gerassimos
  Orkoulas.
\newblock Modeling and control of shape distribution of protein crystal
  aggregates.
\newblock {\em Chemical Engineering Science}, 104:484--497, 2013.

\bibitem{kwon2014crystal}
Joseph~S Kwon, Michael Nayhouse, Gerassimos Orkoulas, and Panagiotis~D
  Christofides.
\newblock Crystal shape and size control using a plug flow crystallization
  configuration.
\newblock {\em Chemical Engineering Science}, 119:30--39, 2014.

\bibitem{kwon2014enhancing}
Joseph~S Kwon, Michael Nayhouse, Gerassimos Orkoulas, and Panagiotis~D
  Christofides.
\newblock Enhancing the crystal production rate and reducing polydispersity in
  continuous protein crystallization.
\newblock {\em Industrial \& Engineering Chemistry Research},
  53(40):15538--15548, 2014.

\bibitem{choi2019modeling}
Hyun-Kyu Choi and Joseph~Sang Kwon.
\newblock Modeling and control of cell wall thickness in batch delignification.
\newblock {\em Computers \& Chemical Engineering}, 128:512--523, 2019.

\bibitem{choi2020multiscale}
Hyun-Kyu Choi and Joseph~S Kwon.
\newblock Multiscale modeling and predictive control of cellulose accessibility
  in alkaline pretreatment for enhanced glucose yield.
\newblock {\em Fuel}, 280:118546, 2020.

\end{thebibliography}
\end{document}